\documentclass[sigconf]{acmart}
\usepackage{supertabular}
\usepackage{wrapfig}
\usepackage{multirow}
\usepackage{array,hhline}
\usepackage{graphicx}
\usepackage{framed}
\usepackage{algorithm,algpseudocode,listings}
\usepackage{xcolor}
\usepackage{caption}
\usepackage{enumitem}
\usepackage{supertabular}
\usepackage{colortbl}
\usepackage[utf8]{inputenc} % allow utf-8 input
\usepackage[T1]{fontenc}    % use 8-bit T1 fonts
\usepackage{microtype}      % microtypography
\usepackage{xcolor}         % colors
\usepackage{xspace}
\definecolor{Light}{rgb}{0.99, 0.92, 0.95}
\definecolor{deemph}{gray}{0.6}
\newcommand{\gc}[1]{\textcolor{deemph}{#1}}
\newcommand{\up}[1]{\textcolor{upcolor}{$\uparrow$ #1}}
\definecolor{textpurple}{RGB}{135,89,201}
\definecolor{upcolor}{RGB}{57,182,74}
\newcommand{\down}[1]{\textcolor{red}{$\downarrow$ #1}}
\newcommand{\pub}[1]{\scriptsize\textcolor{deemph}{[#1]}}

\definecolor{codegreen}{rgb}{0,0.6,0}
\definecolor{codegray}{rgb}{0.5,0.5,0.5}
\definecolor{codepurple}{rgb}{0.58,0,0.82}
\definecolor{backcolour}{rgb}{1.0,1.0,1.0}
\lstdefinestyle{mystyle}{
    backgroundcolor=\color{backcolour},
    commentstyle=\color{codegreen},
    keywordstyle=\color{magenta},
    numberstyle=\tiny\color{codegray},
    stringstyle=\color{codepurple},
    basicstyle=\ttfamily\footnotesize\fontfamily{pcr}\selectfont,
    breakatwhitespace=false,
    breaklines=true,
    captionpos=b,
    keepspaces=true,
    showspaces=false,
    showstringspaces=false,
    showtabs=false,
    tabsize=2
}
\AtBeginDocument{%
  }

\setcopyright{acmcopyright}
\copyrightyear{2023}
\acmYear{2023}
\acmDOI{XXXXXXX.XXXXXXX}

\acmPrice{15.00}
\acmISBN{978-1-4503-XXXX-X/18/06}

\acmSubmissionID{183}

\begin{document}

% \title{ContentCTR: Streaming Highlight Click-Through Rate Prediction with Multi-Modal Transformer}
% \title{ContentCTR: Frame-level Click-Through Rate Prediction with a Multimodal Transformer for Live Streaming Recommendation}
% \title{ContentCTR: Frame-level CTR Prediction with a Multimodal Transformer for Live Streaming Recommendation}
\title{ContentCTR: Frame-level Live Streaming Click-Through Rate Prediction with Multimodal Transformer}
%ContentCTR: Frame-level Prediction with a Multimodal Transformer for Live Streaming Recommendation

\author{Jiaxin Deng}
\authornote{Interns at MMU, KuaiShou Inc.}
\affiliation{%
  \institution{National Laboratory of Pattern Recognition, Institute of Automation, Chinese Academy of Sciences}
  \streetaddress{95 Zhongguancun East Road}
  \city{Haidian Qu}
  \state{Beijing Shi}
  \country{China}
  \postcode{100190}
}
\email{dengjiaxin2022@ia.ac.cn}

\author{Dong Shen}
\affiliation{%
  \institution{KuaiShou Inc.}
  \streetaddress{6 Shangdi West Road}
  \city{Haidian Qu}
  \state{Beijing Shi}
  \country{China}
  \postcode{100085}
}
\email{shendong@kuaishou.com}

\author{Shiyao Wang}
\affiliation{%
  \institution{KuaiShou Inc.}
  \streetaddress{6 Shangdi West Road}
  \city{Haidian Qu}
  \state{Beijing Shi}
  \country{China}
  \postcode{100085}
}
\email{wangshiyao08@kuaishou.com}

\author{Xiangyu Wu}
\affiliation{%
  \institution{KuaiShou Inc.}
  \streetaddress{6 Shangdi West Road}
  \city{Haidian Qu}
  \state{Beijing Shi}
  \country{China}
  \postcode{100085}
}
\email{wuxiangyu@kuaishou.com}

\author{Fan Yang}
\affiliation{%
  \institution{KuaiShou Inc.}
  \streetaddress{6 Shangdi West Road}
  \city{Haidian Qu}
  \state{Beijing Shi}
  \country{China}
  \postcode{100085}
}
\email{yangfan@kuaishou.com}

\author{Guorui Zhou}
\affiliation{%
  \institution{KuaiShou Inc.}
  \streetaddress{6 Shangdi West Road}
  \city{Haidian Qu}
  \state{Beijing Shi}
  \country{China}
  \postcode{100085}
}
\email{zhouguorui@kuaishou.com}

\author{Gaofeng Meng}
\authornote{Corresponding author.}
\affiliation{%
  \institution{National Laboratory of Pattern Recognition, Institute of Automation, Chinese Academy of Sciences}
  \streetaddress{95 Zhongguancun East Road}
  \city{Haidian Qu}
  \state{Beijing Shi}
  \country{China}
  \postcode{100190}
}
\email{gfmeng@nlpr.ia.ac.cn}

\renewcommand{\shortauthors}{Deng et al.}

\begin{abstract}
In recent years, live streaming platforms have gained immense popularity as they allow users to broadcast their videos and interact in real-time with hosts and peers. Due to the dynamic changes of live content, accurate recommendation models are crucial for enhancing user experience. However, most previous works treat the live as a whole item and explore the Click-through-Rate (CTR) prediction framework on \textit{item-level}, neglecting that the dynamic changes that occur even within the same live room. In this paper, we proposed a ContentCTR model that leverages multimodal transformer for \textit{frame-level} CTR prediction. First, we present an end-to-end framework that can make full use of multimodal information, including visual frames, audio, and comments, to identify the most attractive live frames. Second, to prevent the model from collapsing into a mediocre solution, a novel pairwise loss function with first-order difference constraints is proposed to utilize the contrastive information existing in the highlight and non-highlight frames. Additionally, we design a temporal text-video alignment module based on Dynamic Time Warping to eliminate noise caused by the ambiguity and non-sequential alignment of visual and textual information. We conduct extensive experiments on both real-world scenarios and public datasets, and our ContentCTR model outperforms traditional recommendation models in capturing real-time content changes. Moreover, we deploy the proposed method on our company platform, and the results of online A/B testing further validate its practical significance.

\end{abstract}




\maketitle

\section{Introduction}
\begin{figure}
\centering
\includegraphics[width=.45\textwidth]{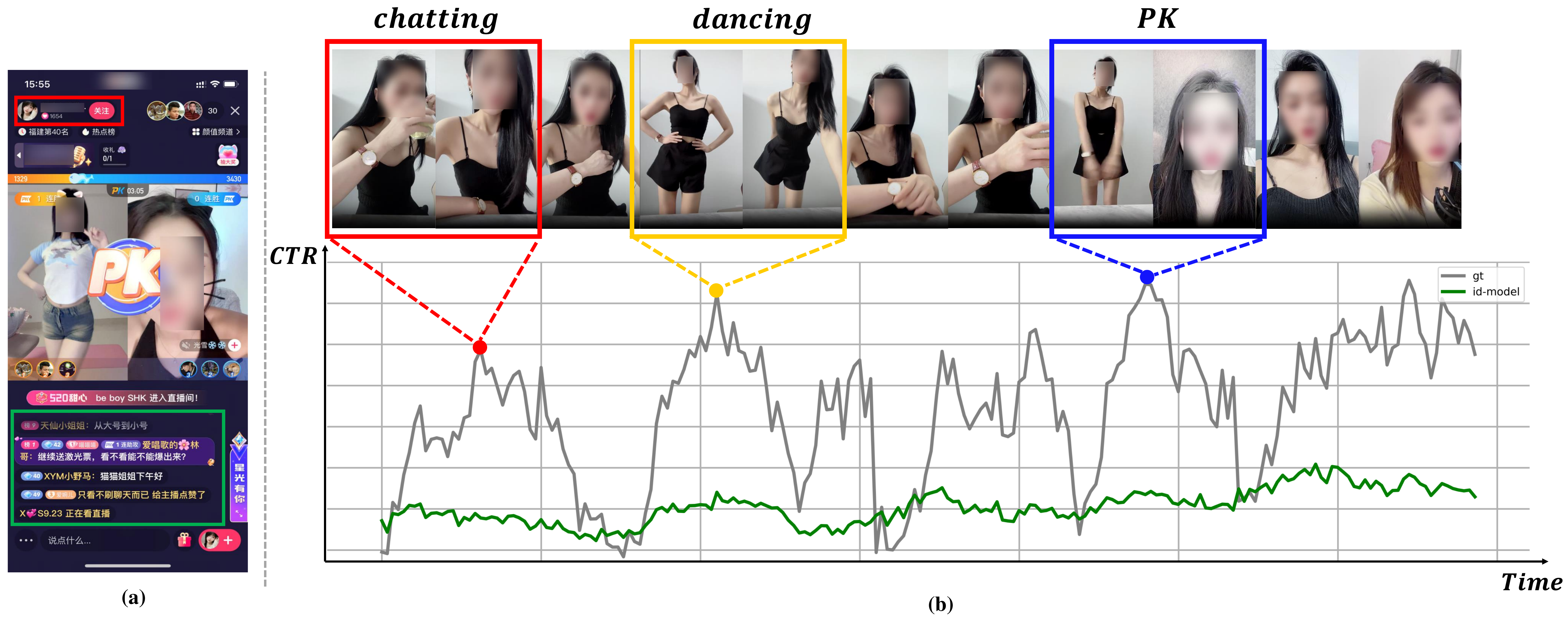}
\caption{\textbf{(a) shows the screenshots of the platform mobile app. There exist four kinds of modality: the visual feature of streaming frames, the speech from streamer, the ID embedding of streamer (red box) and the bullet comments from the user (green box). (b) The grey line shows the ground truth CTR changes of a dancing streamer while the green line represents the predicted CTR of ID-based recommendation model. And three distinct highlight moments are marked by red, yellow, and blue points.
}}
\label{fig1}
\vspace{-0.5cm}
\end{figure}
%As an emerging type of content on multi-modal information-sharing platforms such as TikTok and Kwai, live streaming presents a new arena for competition in the e-commerce market. On such platform, more and more user are attracted by the live-streaming for entertainment while streamers also get the reward from the audience and platform. As shown in Figure \ref{fig1}, streamers present various talent shows on the platform, e.g. video game streaming, chatting and opinion sharing, dancing or singing, PK between streamers and advertisements support by the sponsor and product provider. However, most of the streaming content is tedious and users have limited time to pay attention to the recommendation steaming-room. Finding the most captivating frames from the streaming and showing them to potential interested users have following critical advantages. Firstly, providing users with high-quality streaming would significantly increase user's experience and stickiness towards the platform. Second, captiving steam frames would attract users' interests on steamer's talent and inspire users‘ consumption behavior like paying reward and sending virtual gift to steamer. Thirdly, live-streaming provide a vast market for e-commerce which would create more revenue and incentives for the platform. Therefore, there is a significant demand for the platform to capture and recommend highlighted streaming frames to enhance its user satisfaction, involvement, and loyalty.
Live streaming platform represent a new type of online interaction and experienced rapid growth in recent years. On such platform, more and more users are attracted to live streaming for entertainment, while streamers receive rewards from both the audiences and the platform. This new form of interaction and entertainment has motivated researchers to study emerging issues such as gift-sending mechanisms \cite{zhu2017understanding}, E-Commerce events \cite{yu2021leveraging}, and other live streaming practices. Due to the large number of streamers and the constantly changing content, an accurate recommendation algorithm is crucial for enhancing user experience. \cite{zhang2021deepsequential,zhang2021deep1} focus on capturing mutual information between streamers and viewers to obtain better representations. \cite{rappaz2021recommendation} propose a self-attentive model to address the dynamic availability and repeat consumption of live streaming. \cite{yu2021leveraging} present a novel Live Stream E-Commerce Graph Neural Network framework to learn the tripartite interaction among streamers, users and products.

%Traditional item-level video recommendation works \cite{tao2022self,yi2022multi,he2021click,wei2021contrastive} incorporates multi-modality feature for video-level Click-Through Rate (CTR) predication. However, under the circumstance of live-streaming highlight recommendation, the frames level CTR prediction architecture is need. Thus, this make stream highlight recommendation more challenging than traditional video recommendation task. In order to frame-level CTR accurately, five challenges should be concerned: 1). As shown in Figure \ref{fig1}, there exists four kinds of modality in live-streaming scenerio: the visual feature of streaming frames, the speech from streamer, the ID embedding of streamer and the bullet comments from the user.  Therefore, a model can  leverage different modality is essential. (2). A frame level CTR prediction network is need for captiving highlight streaming frames. (3). Model the contrastive relationship among the highlight frames and no-highlight frames is beneficial to generate the ﬁnal CTR prediction results. (4). Due to the consecutiveness of live streaming, a model which is capable for onling learning to follow the dynamic patterns based on the history content of live-streaming is needed. (5). In the scenario of live-streaming, the speech of streamer and the comment from the user maybe be ambiguous and non-sequential alignment to the visual frames, so a module is also need to eliminate the noise from the mis-alignment.
The broadcast of real-time content implies the live is constantly changing, and the performance of various segments may differ significantly. Although most previous work \cite{zhang2021deepsequential,zhang2021deep1,yu2021leveraging,rappaz2021recommendation} treat live broadcasts as items and model them at the item-level, we believe that understanding the content of individual frames and making more refined, frame-level estimates is crucial. Additionally, identifying the most captivating frames from the stream and presenting them to potentially interested users can considerably enhance user engagement, attract more users, and increase revenue for live-streaming platforms. To fully capture the changes in live content, three challenges must be addressed: (1) As depicted in Figure \ref{fig1} (a), live-streaming scenarios involve several modalities, including real-time visual frames, speech from the streamer, bullet comments from audiences, and other categorical inputs (e.g., streamer IDs and live IDs). Consequently, a model that can leverage various modalities and their historical information is crucial. (2) Traditional pointwise estimation may lead to the model converging on a suboptimal solution. Apart from that, as illustrated in Figure \ref{fig1} (b), the traditional ID based model may learn a mediocre solution (shown in green line), neglecting multiple singular points that could represent live streaming highlights. Thus, an appropriate loss function that models the contrastive relationship between highlight and non-highlight frames is beneficial for generating the final CTR prediction results. (3) In live-streaming scenarios, the streamer's speech and audiences' comments may be ambiguous and not sequentially aligned with the visual frames, necessitating a module to filter out the noise caused by misalignment.

In this paper, we propose an end-to-end transformer-based network called ContentCTR, which leverages multi-modality features for frame-level Click-Through Rate (CTR) prediction. To the best of our knowledge, this is the first research that explores frame-level live streaming recommendation. First, the ContentCTR exploits the information contained in visual, speech, comment, and ID embedding and uses an attention mechanism to adaptively find correlations and interactions from different modalities. Second, to alleviate the misalignment between visual and textual modality, we develop a dynamic time warping algorithm to address potential temporal discrepancies that may arise during live streaming events. Finally, to better exploit the contrastive information between highlight frames and no-highlight frames, we design a pairwise loss function with first-order difference constraints. We find that the constraints are essential when jointly optimizing pointwise and pairwise losses to avoid collisions and model collapse. It is worth noting that our work focuses on modeling the relationship between real-time live content and CTR, without introducing user personalization factors. The model can be utilized not only for online traffic mechanism but also for helping existing recommendation models enhance content understanding. We perform comprehensive experiments on both a large-scale real-world live streaming dataset and a public PHD dataset \cite{garcia2018phd} and achieve the state-of-the-art performance. 

In summary, the main contributions made in this work are as follows:

\begin{itemize}[leftmargin=*]
\item We propose ContentCTR, an end-to-end transformer-based network for frame-level CTR prediction in live streaming platform. It efficiently utilizes the features of different modalities and captures the dynamic highlight pattern.
\item We design a pairwise-based loss function with first-order difference constraints to exploit the contrastive information of highlight frames and no-highlight frames. In addition, a dynamic time warping (DTW) based alignment strategy is present to alleviate misalignment in streaming scenarios. 
\item We conduct comprehensive experiments on both a large-scale real-world live streaming dataset and a public PHD dataset \cite{garcia2018phd}. Our method outperforms all baseline models. We also perform online A/B testing on the live streaming platform of company, achieving a 2.9\% lift in CTR and a 5.9\% improvement in terms of live play duration.
\end{itemize}

\begin{figure*}[h]
\centering
\includegraphics[width=.95\textwidth]{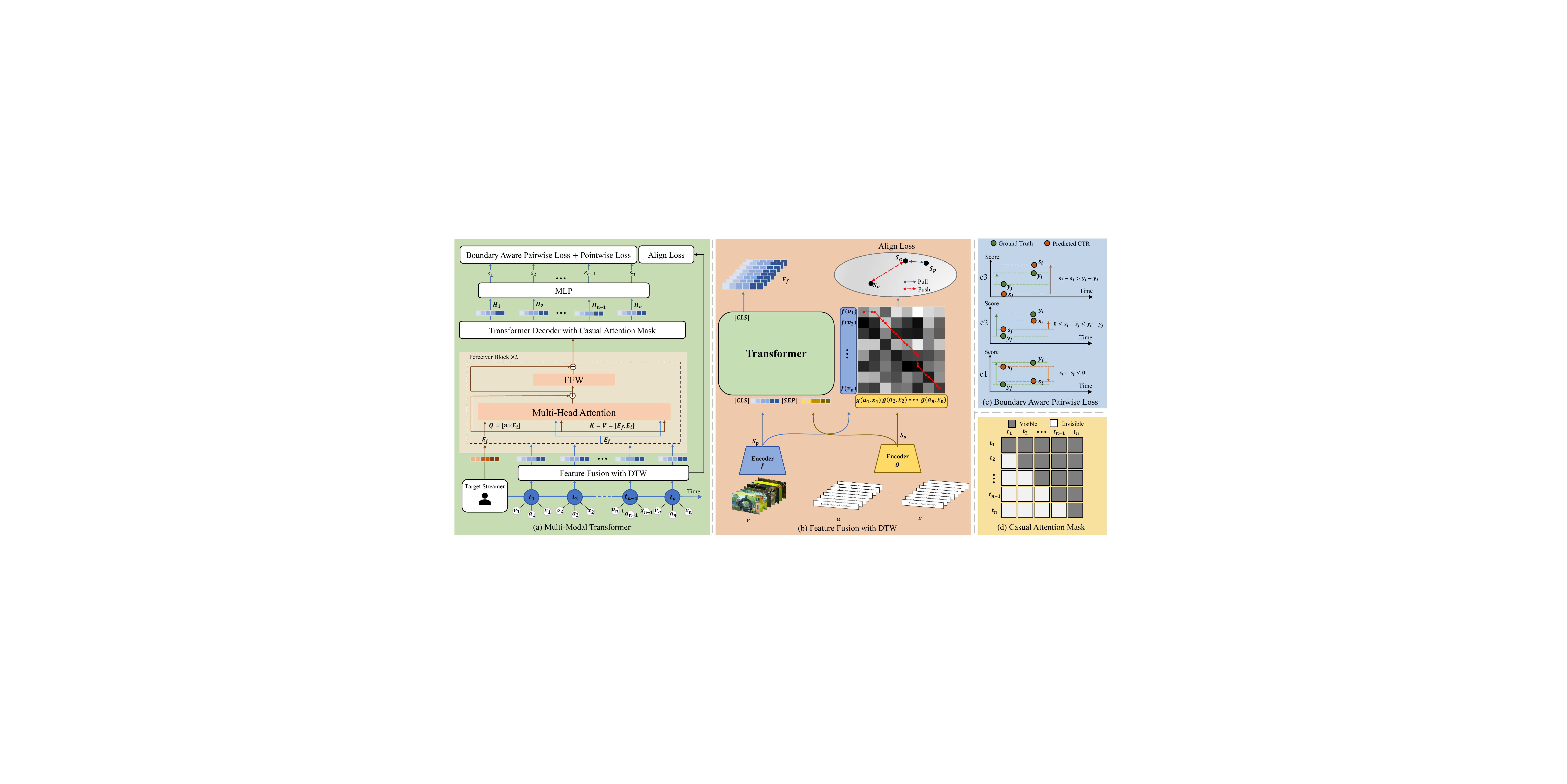}
\caption{\textbf{The framework of our proposed method.} Part (a) and (d) show the architecture of Multi-Modal Transformer and Casual Attention Mask which are discussed in Section \ref{MultiModalTransformer}. Part (b) shows the proposed Dynamic Time Warping Alignment Loss which is discussed in Section \ref{DTWAlignment}. Part (c) shows the motivation of Boundary-aware Pairwise Loss which is discussed in Section \ref{BoundaryAwarePairwiseLoss}.}
\label{fig2}
\vspace{-0.4cm}
\end{figure*}

\section{Related Work}
%参考Enhancing CTR Prediction with Context-Aware Feature Representation Learning写一写比较笼统的CTR模型的related work
% Existing works on video CTR prediciton can be divided into two main categoroes, i.e., Factorization Machine (FM) \cite{rendle2010factorization} based methods and deep nerual networks (DNNs) based methods.  For FM based method, 
%Wangshy：related work都没有介绍直播推荐相关的工作呀？要补上，比如
%A deep bi-directional prediction model for live streaming recommendation
%A Hybrid Preference-Aware Recommendation Algorithm for Live Streaming Channels
%Deep_sequential_model_for_anchor_recommendation_on_live_streaming_platforms
%Live streaming recommendations based on dynamic representation learning
%Recommendation on Live-Streaming Platforms- Dynamic Availability and Repeat Consumption
% As the “interesting” video frames are subjective to the viewer, most video highlight methods do not consider user preference \cite{gygli2016video2gif,jiao2018video,yao2016highlight,yu2018deep,hong2020mini,badamdorj2021joint,xu2021cross,wei2022learning,wang2023hard} and these works typically optimize variants of a pairwise ranking loss, ranking the highlight frames higher than the non-highlight frames.
\subsection{Deep Click-Through Rate Prediction}
 Based on the input modality, the existing methods for Click-Through Rate (CTR) prediction can be classified into two categories: ID-based and multi-modal-based CTR prediction models. For the ID-based CTR prediction approachs, techniques such as Wide\&Deep \cite{cheng2016wide}, DeepFM \cite{guo2017deepfm}, and DCN-M \cite{wang2021dcn} aim to capture high-order feature interactions and complex feature dependencies. On the other hand, DIFM \cite{lu2021dual} addresses the fixed feature representation problem by learning the input vector level weights for feature representations. For multi-modal-based CTR prediction models, the majority of the research has focused on video Click-Through Rates prediction, which is a fundamental task in the field of multimedia and information retrieval. For example, AutoFIS \cite{liu2020autofis} propose a two-stage algorithm called Automatic Feature Interaction Selection while UBR4CTR \cite{qin2020user} propose User Behavior Retrieval for CTR prediction framework to tackle the problem that sequential patterns such as periodicity or long-term dependency are not embedded in the recent several behaviors but in far back history. Both of these approaches are based on the Factorization Machine (FM) \cite{rendle2010factorization} method. For DNNs based methods, DSTN \cite{ouyang2019deep} propose a Deep Spatiotemporal Translation Network (DSTN), which utilize the auxiliary data in spatial and temporal domain. In order to better capture user preferences, HyperCTR \cite{he2021click} design a Hypergraph Click-Through Rate prediction framework, built upon the hyperedge notion of hypergraph neural networks, which can yield modal-specific representations of users and micro-videos. However, all these methods only focus on video-level CTR prediction. Compared with traditional video CTR prediction task, stream highlight CTR prediction needs frames level CTR prediction network which can leverage multi-modal feature to address its unique challenges, i.e. the dynamic patterns of live-streaming and non-sequential alignment of different modality.
\subsection{Live Streaming Recommendation}
As an emerging form of social media, increasing attention has being given to live streaming recommendations. Current works typically treat streamers or audiences as items, and develop recommendation systems to predict interactions between users and items. These methods can be broadly categorized into two types: Filtering-based methods and Deep learning-based methods. For Filtering-based methods, HyPAR \cite{yang2013hybrid} proposed a Collaborative and Content-based filtering hybrid approach to recommend streamers to audiences. For Deep learning-based methods, LiveRec \cite{rappaz2021recommendation} and \cite{zhang2021deep1} leverages self-attentive mechanisms based on historical interactions and repeat consumption behavior, while DRIVER \cite{gao2023live} learns dynamic representations by leveraging users’ highly dynamic behaviors. \cite{zhang2021deep} utilizes LSTM to extract the representation and capture the preference of streamers and audiences. However, these works only focus on the interaction between streamers and audiences at the item level, neglecting the importance of highly dynamic multi-modal content in live streaming. Additionally, they cannot provide frame-level recommendations.
\subsection{Video Highlight Detection}
The task most closely related to stream CTR prediction is Personalized Video Highlight Detection (P-VHD), as both aim to recognize different content patterns in temporal sequences. The primary objective of P-VHD is retrieving a subset of video frames that capture a person's main attention or interests from the original video. Recently, several works \cite{garcia2018phd,rochan2020adaptive,chen2021pr,wang2022pac,bhattacharya2022show} are proposed for P-VHD task, which aims at extracting user-adaptive highlight predictions guided by annotated user history. For example, PHD-GIFs \cite{garcia2018phd} is the first personalized video highlight detection technique that also creates a large-scale dataset called PHD. The P-VHD task and stream CTR prediction both need to consider personalized behavior, while P-VHD only needs to predict the label of frames to highlight or no-highlight one. On the other hand, stream CTR prediction needs to regress the correct order of all frames, which means that stream highlight CTR prediction is more challenging.

\section{Methodology}
As illustrated in Figure \ref{fig2}, we propose the Transformer Click-Through Rate prediction framework, named ContentCTR. ContentCTR utilizes multi-modal features to predict the CTR at frame level. In particular, we first define the problem of predicting CTR for stream highlights in Section \ref{problem}. Then the Multi-Modal Transformer backbone is introduced to fuse and interact with different modalities in Section \ref{MultiModalTransformer}. Additionally, we propose the Dynamic Time Warping strategy for aligning text and visual feature in Section \ref{DTWAlignment}. Finally, boundary-aware pairwise loss for exploring contrasting information is presented in Section \ref{BoundaryAwarePairwiseLoss}.
\subsection{Problem Formulation}
\label{problem}
The main goal of the CTR prediction is to estimate the probability of the current content of the live broadcast being clicked. In this work, we simplify this problem for predicting the Click-Through Rate of frame $\delta_i$ at timestamp $i$. We denote $M_i=\{v_i, a_i, x_i\}$ as the multi-modal tuple, where $v_i,a_i$ and $x_i$ represent the visual, speech and comments of frame $\delta_i$, respectively. We hypothesize that different streamers have distinct talents and attract different audiences who are typically interested in specific types of highlight moments in a streaming room. For instance, as shown in Figure \ref{fig1} (b), some audiences may be attracted to a streamer's dancing, while others may be attracted to the PK between streamers. Therefore, we denote $E_u$ as the ID embedding of streamer $u$. The ID embedding of streamer is extracted by a pre-trained SIM \cite{pi2020search} model that reflects the rough classification of a streamer. Additionally, we consider that the highlight pattern and audience taste change over time, and thus, the model should use information from the $n-1$ lookahead windows $W_M=\{M_{i-n+1},M_{i-n+2},\cdots, M_{i-1}\}$ of previous frames to predict the CTR $y_i$ of frame $\delta_i$. The prediction problem can be formulated as follows,
\begin{equation}
\operatorname{Prob}\left(\delta_i \mid W_M, E_u \right) \sim \Gamma\left( W_M, E_u, \delta_i \right)
\end{equation}
where the frame $\delta_i$ is represented by multi-modal feature of lookahead windows $W_M$ and the ID embedding $E_u$. The probability that audiences will click on the frame $\delta_i$ is denoted by $\operatorname{Prob}\left(\delta_i \mid W_M, E_u \right)$. Moreover, $\Gamma\left( W_M, E_u, \delta_i \right)$ is the model used to estimate the probability $\operatorname{Prob}\left(\delta_i \mid W_M, E_u \right)$.

\subsection{Multi-Modal Transformer}\label{MultiModalTransformer}
The proposed Multi-Modal Transformer backbone includes the following three basic components, i.e., feature fusion layer, Perceiver block, and casual sequence decoder.

%公式参考mPLUG-2: A Modularized Multi-modal Foundation Model  Across Text, Image and Video
\subsubsection{Feature Fusion Layer}\label{FeatureFusionLayer} As depicted in Figure \ref{fig2} (b), given the historical window $W_M$ of streamer $u$, we extract multi-modal features for every timestamp, including the streaming frames $v_i$, Auto Speech Recognition (ASR) text $a_i$ from the streamer, comment $x_i$ from the audiences. The streaming frames are tokenized by the pre-trained swin \cite{liu2021Swin} $f(\cdot)$ while the ASR and comment text are tokenized with the BERT \cite{devlin2018bert} Chinese Large model $g(\cdot)$ and we only use the hidden feature of \texttt{<CLS>} token as the text embedding. Since the embedding dimensions for streaming frames and language tokens are different, two MLP heads $proj_1(\cdot)$ and $proj_2(\cdot)$ are set to map the frames embedding and text embedding to the same dimension $d$, denoted by $S_p  \in \mathcal{R}^{b \times n \times 1 \times d} $ and $S_a  \in \mathcal{R}^{b \times n \times 1 \times d}$, where $n$ is the length of lookahead window and $b$ is the batch size. The above process can be formulated as follows:
\begin{equation}
\begin{aligned}
{S}_{{p}} & ={proj}_1(f(\boldsymbol{v})) \\
{S}_{{a}} & ={proj}_2\left({g}_{\texttt{<{CLS}>}}([\boldsymbol{a}, \boldsymbol{x}])\right)
\end{aligned}
\end{equation}
where $\boldsymbol{v}=\left[ v_1,\cdots ,v_n \right] , \boldsymbol{a}=\left[ a_1,\cdots ,a_n \right]$ and $ \boldsymbol{x}=\left[ x_1,\cdots ,x_n \right] $.

For frames $\delta_i$ at timestamp $i$, the corresponding frames feature and text feature are denoted by $S_{p}^{i} \in \mathcal{R}^{b \times 1 \times 1 \times d}$ and $S_{a}^{i} \in \mathcal{R}^{b \times 1 \times 1 \times d}$ respectively. We concatenate $S_{p}^{i}$ and $S_{a}^{i}$ as the final input tokens $E_{f}^{i} \in \mathcal{R}^{b \times 1 \times  2 \times d} $ of the Perceiver block. Finally, we obtain $E_f \in \mathcal{R}^{b \times n \times  2 \times d} $ by concatenating all features of the timestamps.
\subsubsection{Perceiver Block}\label{PerceiverBlock}
The Perceiver block is connected to the casual sequence decoder, as shown in Figure \ref{fig2} (a). It takes the repeated ID embedding $n \times E_u \in \mathcal{R}^{b \times n \times 1 \times d}$ of streamer $u$ and the fusion multi-modal feature $E_f \in \mathcal{R}^{b \times n \times 2 \times d}$ as input, where $n$ is the length of lookahead window, $b$ is the batch size and $d$ is the input dimension. The motivation of Perceiver block is the success of multi-head attention mechanism \cite{vaswani2017attention} in sequential recommendation \cite{chen2019behavior} and video understanding \cite{alayrac2022flamingo}. We apply it to capture streamer's highlight patterns on lookahead sequences with the query-based retrieval problem. 

First, we initialize learned latent query $Q \in \mathcal{R}^{bn \times 1 \times d}$ with flattened $n \times E_u$. Next, we concatenate the flattened $E_f$ and $n \times E_u$ at the second dimension and take $K = V = [E_f, n \times E_u] \in \mathcal{R}^{bn \times 3 \times d}$ as key and value. We linearly project the input vector to latent vectors with $d_h$ dimensions, by different linear projections. Then we perform the scaled dot-product multi-head attention as follows,
\begin{equation}
\begin{aligned}
\operatorname{MultiHead}(Q, K, V) & =\operatorname{Concat}\left(\text {head}_1, \text{head}_2, \cdots, \text{head}_{n_h}\right), \\
\text{head}_i & =\operatorname{Attention}\left(Q W_i^Q, K W_i^K, V W_i^V\right),
\end{aligned}
\label{atten1}
\end{equation}
where $n_h$ is the number of attention head and $W_i^Q \in \mathcal{R}^{d \times d_h}, W_i^K \in \mathcal{R}^{d \times d_h}$ and $W_i^V \in \mathcal{R}^{d \times d_h}$ are the learnable parameters. The scaled dot-product attention function is defined as follows,
\begin{equation}
\operatorname{Attention}(Q, K, V)=\operatorname{softmax}\left(\frac{Q K^{\top}}{\sqrt{d_h}}\right)
\label{atten2}
\end{equation}
Then, the Perceiver block employs a feed-forward network with residual connections for performance boosting. The Perceiver block is stacked for $L$ layers and the outputs of previous block are fed into the Perceiver block alternately. The output of Perceiver block is denoted by $E_p \in \mathcal{R}^{bn \times 1 \times d_n}$. The pseudocode of the Perceiver block is shown in Algorithm \ref{alg1}.
\begin{algorithm}[h]
\caption{A Pytorch-style Pseudocode for Perceiver Block.}
\label{alg1}
\begin{lstlisting}[language=python]
def perceiver_block(
    x_f,  # The [b, n, 2, d] multi-modal feature
    x, # The learned latent query with shape [b, n, 1, d]
    num_layers, # The number of layers
): 
    x_f = flatten(x_f) # [b, n, 2, d] -> [b * n, 2, d]
    x = flatten(x) # [b, n, 1, d] -> [b * n, 1, d]
    for i in range(num_layers):
        # Attention
        x = x + attention_i(q=x, kv=concat([x_f, x]))
        # Feed forward with residual connection
        x = x + ffw_i(x)
    return x
\end{lstlisting}
\end{algorithm}
% \vspace{-0.5cm}
\subsubsection{Casual Sequence Decoder}\label{CasualSequenceDecoder}
We first unflatten and squeeze the output of Perceiver block $E_p \in \mathcal{R}^{bn \times 1 \times d}$ as the dimension of $\mathcal{R}^{b \times n \times d}$. Then we initialize the query, key and value of decoder as $Q = K = V = E_p \in \mathcal{R}^{b \times n \times d}$. After that, we apply the scaled dot-product multi-head attention which is the same as shown in Equation \ref{atten1}, but with the scaled dot-product attention function and casual attention as follows:
\begin{equation}
\operatorname{Attention}(Q, K, V)=\operatorname{softmax}\left(\frac{Q K^{\top}}{\sqrt{d_h}} + M \right)
\label{atten3}
\end{equation}
where $M$ is the casual attention mask. As shown in Figure \ref{fig2} (d), it is an $n \times n$ matrix filled with \text{-inf} and its upper triangular sub-matrices are filled with 0. By applying the casual attention mask $M$ the problem of future information leakage in the temporal dimension is avoided. The pseudocode of the Decoder is shown in Algorithm \ref{alg2}.
\begin{algorithm}[h]
\caption{A Pytorch-style Pseudocode for Decoder.}
\label{alg2}
\begin{lstlisting}[language=python]
def decoder(
    x, # The input with shape [b*n, 1, d_n]
    num_layers, # The number of layers
): 
    x = unflatten(x).squeeze() #[b*n,1,d_n] -> [b,n,d_n]
    for i in range(num_layers):
        # Attention
        x = x + attention_with_mask_i(q=x, k=x, v=x))
        # Feed forward with residual connection
        x = x + ffw_i(x)
    return x
\end{lstlisting}
\end{algorithm}
The output $H$ from the final attention layer is then fed into a fully-connected layer, followed by a sigmoid transformation to produce the scalar prediction of CTR $s$. The parameters of the fully-connected layer are represented by $W \in \mathcal{R}^{d_n \times 1}$.

The main objective is to maximize the log-likelihood between the predicted CTR $s$ and the actual CTR $y$, which is achieved using the following pointwise model to optimize the standard LogLoss \cite{liu2009learning}:
% \begin{equation} 
% % L_{CE} = -\frac{1}{n}\sum_{i=1}^{n}\left[y_i\cdot\log\left(s_i\right) + (1-y_i)\cdot\log\left(1-s_i\right)\right]
% L_{List}=-\sum_{i=1}^n{\frac{y_i}{\sum_{j=1}^n{y_j}}}\log\mathrm{(}\frac{e^{s_i}}{\sum_{j=1}^n{e^{s_j}}})
% \end{equation}

\begin{equation} 
L_{Point} = -\frac{1}{n}\sum_{i=1}^{n}\left[y_i\cdot\log\left(s_i\right) + (1-y_i)\cdot\log\left(1-s_i\right)\right]
\end{equation}

\subsection{DTW Alignment}\label{DTWAlignment}
The motivation behind for alignment is to address potential temporal discrepancies that may arise during live streaming events. For example, the streamer may describe the plan before taking action, or explain detailed information after action. Additionally, the comments from the audiences may experience some time lag, which further exacerbates the misalignment issue. Therefore, it is essential to train text and visual encoders that can handle misalignment to alleviate that problem. Inspired by previous works \cite{xu2021videoclip,ko2022video}, in this section we present our contrastive learning based framework for visual and text sequences alignment.

Given the text sequence $S_a$ and visual sequence $S_p$ from a lookahead window $W_M$, we assume that when timestamps $i$ in the text sequence and $j$ in the visual sequence represent the same pattern, then $S_a^{i}$ and $S_p^{j}$ should have a semantic similarity. To train a network that can minimized the distance between text sequences $S_a$ and visual sequences $S_p$, we use the Dynamic Time Warping \cite{muller2007dynamic} (DTW) which calculates the minimum cumulative matching costs over units as the sequence distance to measure the similarity. 
First, we computes a pairwise distance matrix $D\left( S_a,S_p \right) \coloneqq \left[ \mathrm{sim} \left( S_{a}^{i},S_{p}^{j} \right) \right] _{ij}\in \mathcal{R} ^{n\times n}$ with a distance measure $\mathrm{sim}(\cdot)$. In our work we applied the cosine similarity as $\mathrm{sim}(\cdot)$.
Then, we employs dynamic programming and sets a matrix $C \in \mathcal{R}^{n \times n}$ to record the minimum cumulative cost between $S_a^i$ and $S_p^j$ \cite{dixit1990optimization}:
\begin{equation}
C_{i,j} =D_{i,j} +\min \left\{ C_{i-1,j-1} ,C_{i-1,j} ,C_{i,j - 1} \right\} 
\label{dtwdp}
\end{equation}
where $1\le i,j\le n$. Then, the distance $d_{\left\{\mathbf{S}_a, \mathbf{S}_p\right\}}$ between sequences $S_a$ and $S_p$ is set to the last element of matrix $C$:
\begin{equation}
d_{\left\{\mathbf{S}_a, \mathbf{S}_p\right\}}=C_{n,n}
\label{dtwdpdis}
\end{equation}

Contrastive Learning (CL) is a self-supervised learning technique that learns a representation of data by comparing different views of same samples. The basic idea of CL is to bring together similar pairs and push away dissimilar pairs. We hypothesize that the positive pair $\left\{{S}_a, {S}_p\right\}$ should be similar. To construct the negative sample ${S}_n$, we randomly shuffle the temporal order of video sequence ${S}_p$. Therefore, the negative pair $\left\{{S}_a, {S}_n\right\}$ should be dissimilar. Then, we can derive the training objective to minimize the InfoNCE loss \cite{oord2018representation}:
{\footnotesize
\begin{equation}
\begin{aligned}
\mathcal{L} _{align}\left( S_a,S_p,\boldsymbol{S}_n \right) = -\log \frac{\exp \left( d_{\left\{ S_a,S_p \right\}}/\tau \right)}{\exp \left( d_{\left\{ S_a,S_p \right\}}/\tau \right) +\sum_{S_n\in \boldsymbol{S}_n}{\exp}\left( d_{\left\{ S_a,S_n \right\}}/\tau \right)}
\end{aligned}
\end{equation}
}
where $\boldsymbol{S}_n=\left\{S_{ni}\right\}_{i=1}^N$ is the set of $N$ negative samples shuffled from $S_p$. As shown in Figure \ref{fig2} (b), by minimizing the align loss $\mathcal{L} _{align}$, the visual encoder $f(\cdot)$ and text encoder $g(\cdot)$ should be able to learn a good representation from aligned and misaligned pairs.

\subsection{Boundary Aware Pairwise Loss}\label{BoundaryAwarePairwiseLoss}
\begin{figure}[h]
\centering
\includegraphics[width=.45\textwidth]{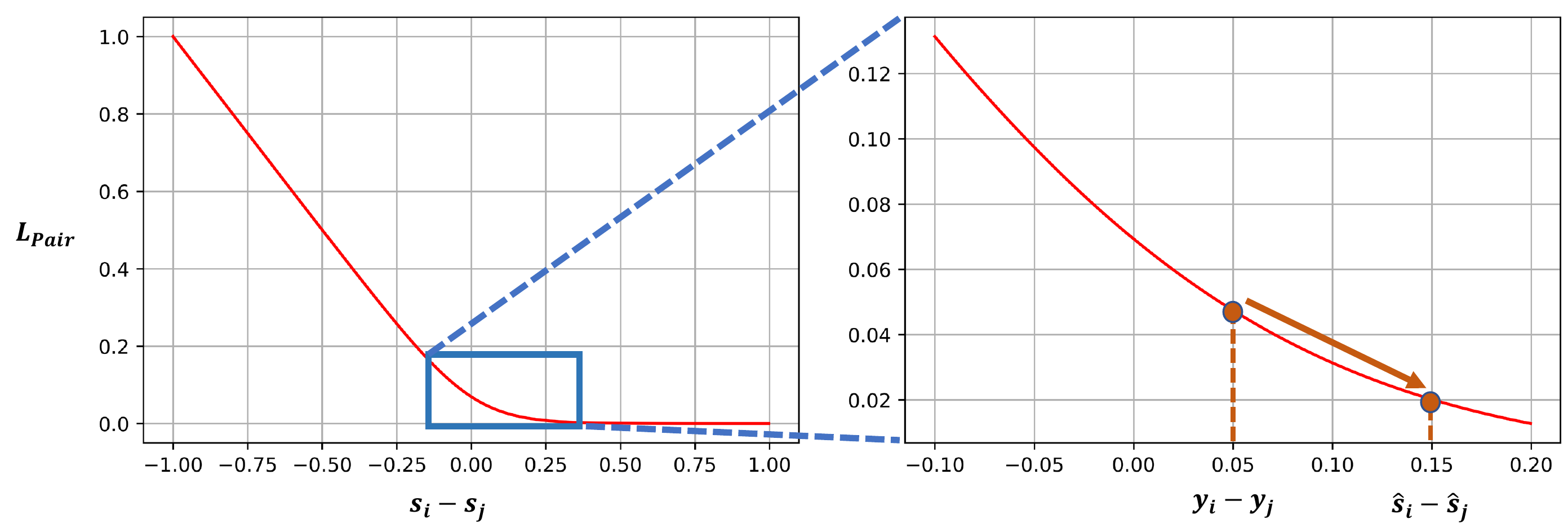}
\caption{\textbf{Loss function for pairwise optimization without any constraints.}}
\label{fig3}
\vspace{-0.5cm}
\end{figure}
In this section, we present some intriguing discoveries on the traditional pairwise logistic ranking loss \cite{burges2005learning} and propose modifications to the loss function by integrating the boundary-aware first-order difference constraint, thereby enhancing the optimization. The addition of the pairwise loss aids in exploiting the underlying contrasting information present in both the highlight and no-highlight frames.

Consider the following pairwise loss function, which has no constraints:
\begin{equation}
L_{Pair}^0=\sum_{y_i>y_j}{\log \left( 1+e^{-\sigma\left( s_i-s_j \right)} \right)}
\label{paitlsoos}
\end{equation}
where $y_i$ and $y_j$ represent the ground truth CTR at timestamps $i$ and $j$, while $s_i$ and $s_j$ represent the predicted CTR from the model. Figure \ref{fig3} illustrates the changing of the loss function $L_{Pair}$ with respect to $s_i-s_j$, which reveals that when $y_i-y_j$ holds, minimizing $L_{Pair}$ tends to cause $s_i-s_j$ to overtake the optimal value of $y_i-y_j$. This leads to over-optimization. 

Based on above findings, we propose a revised pairwise loss that incorporates a boundary-aware first-order difference constraint:
\begin{equation}
L_{Pair}^1=\sum_{y_i>y_j}{\log \left( 1+e^{-\sigma \left( s_i-s_j \right)} \right)},\left( y_i-y_j \right) -\left( s_i-s_j \right) \geqslant 0
\end{equation}
where $\left( y_i-y_j \right) -\left( s_i-s_j \right) \geqslant 0 $ denotes the boundary, and solely those samples that reside within the boundary will calculate $L_{Pair}^1$. Any samples located outside the boundary will result in $L_{Pair}^1$ being set to 0.

Without losing generality, the original pairwise loss function $L_{Pair}^0$ presented in Equation \ref{paitlsoos} can be divided into three distinct parts:
\begin{itemize}
\item Part 1: As shown in Figure \ref{fig2} (c1), when $s_i-s_j \leq 0$, the model's assessment of the significance between timestamp $i$ and $j$ is entirely incorrect, given that timestamp $i$ is more "highlighting" than timestamp $j$.
\item Part 2: As shown in Figure \ref{fig2} (c2), when $0 < s_i-s_j < y_i-y_j$, it implies that the model has distinguished that timestamp $i$ is more "highlighting" than timestamp $j$, but it still fails to accurately predict the difference in CTR values between the two timestamps, and thus, it is still not optimal.
\item Part 3: As shown in Figure \ref{fig2} (c3), when $ y_i-y_j < s_i-s_j$, it indicates that the model's predicted CTR value is too aggressive, resulting in over-optimization.
\end{itemize}

In order to verify above scenarios, we design the following loss functions:
\begin{equation}
L_{Pair}^{2}=\sum_{y_i>y_j}{\log \left( 1+e^{-\sigma \left( s_i-s_j \right)} \right)},s_i-s_j \leq 0
\end{equation}
where $s_i-s_j\leq0$ means that it only optimizes on Part 1.

\begin{equation}
\begin{aligned}
L_{Pair}^{3}=\sum_{y_i>y_j}{\log \left( 1+e^{-\sigma \left( s_i-s_j \right)} \right)},y_i-y_j>s_i-s_j>0
\end{aligned}
\end{equation}

where $y_i-y_j>s_i-s_j>0$ means that it only optimizes on Part 2. The ablation study among  $L_{Pair}^{0}$,  $L_{Pair}^{1}$,  $L_{Pair}^{2}$ and  $L_{Pair}^{3}$ is discuss in Section \ref{exper}. In this work, we apply the $L_{Pair}^{1}$ for optimization.

By combining pointwise loss, align loss and pairwise loss, our final loss used to learn the model parameters is defined as:
\begin{equation}
\mathcal{L}=\lambda_{1}\mathcal{L}_{Point}  + \lambda_{2}\mathcal{L}_{align} +  \lambda_{3}\mathcal{L}_{Pair}^1
\label{multiloss}
\end{equation}
where $\lambda_{1}$, $\lambda_{2}$ and $\lambda_{3}$ are the tradeoff parameters.

\section{Experiments}\label{exper}
In this section, we first introduce the datasets, evaluation metrics, and implementation details. Next, we present the experimental results and some analysis of them. Specifically, our experiments aim to answer the following research questions:
\begin{itemize}
\item \textbf{RQ1}: How does the modality impact the model?
\item \textbf{RQ2}: To what extent can the proposed Perceiver Block enhance the model's performance?
\item \textbf{RQ3}: How effective is the proposed DTW alignment block in alleviating misalignment cases?
\item \textbf{RQ4}: How much can the model's performance be enhanced by the proposed pairwise loss?
\item \textbf{RQ5}: How well does the proposed model perform on public datasets, such as video highlights?
\end{itemize}

\subsection{Dataset}
To comprehensively evaluate the model's performance, we present experimental results on both KLive Dataset and a public video highlighting dataset. Detailed information regarding these datasets is provided as follows.
\subsubsection{KLive Dataset} We have constructed a large-scale dataset based on our company's short video and live streaming platform, which boasts over three million daily active users. Our methodology involved selecting 39,000 high-quality live rooms over a period of three days, based on user rewards and viewing time indicators. Each live room is then divided into multiple consecutive 30s live segments, with three pictures evenly sampled for each segment. The streamer's ASR and audiences' comments are extracted for each segment, and the CTR is calculated by dividing the number of clicked users by all watched users. We then construct consecutive 20 streaming segments into a dataset sample, filtering out samples with a low number of watched users of the last live segment to ensure a reliable CTR. In the end, we obtained 1,436,979 and 286,510 samples for our training and test datasets, respectively. Each sample is comprised of the streamer's item ID, three video frames for 20 streaming segments, audience comments, the streamer's ASR speech, and the ground truth CTR for the all live segments.
\subsubsection{PHD Dataset}
To verify the generality of our method, we evaluate on the publicly available personalized video highlight detection dataset \cite{garcia2018phd} (PHD). This dataset comprises of URLs of YouTube videos, IDs of evaluators or "users," and the segments they designated as highlight frames based on their preferences. The most recent video that a user annotated is considered as the target video for that particular user. Since PHD only provides YouTube URLs, we download the original videos and crawl the captions from YouTube to carry out the experiments. Our training and testing sets include up to 20 history highlight frames and one target video per user. The duration of the history highlight frames varies from 1 to 672.19 seconds, with an average length of 5.12 seconds. The target videos range from 1 to 37,434 seconds, with an average length of 431.79 seconds. The training set comprises of a total of 12,541 users and 81,056 videos, while the testing set has 833 users and 7,595 videos that do not overlap with any user or video in the training set. Consistent with \cite{chen2021pr}, we divide the target video into fixed-length segments (192 frames) and only train those segments that contain highlight frames. During testing, we make the entire video segments as input for inference.

\subsection{Evaluation Metrics}
For the experiments on KLive dataset, we employed Kendall’s tau $\tau$ \cite{kendall1945treatment} to measure the correlation between our predicted click-through rates (CTR) $s$ and the ground truth CTR $y$. It is defined as:
\begin{equation}
\tau =\frac{P-Q}{\sqrt{\left( P+Q+T \right) \cdot \left( P+Q+U \right)}}
\end{equation}
where $P$ represents the number of concordant pairs, $Q$ denotes the number of discordant pairs, $T$ indicates the number of ties only in $s$, and $U$ is the number of ties only in $y$. If the same pair experiences a tie in both $s$ and $y$, it is not included in either $T$ or $U$. Value of $\tau$ close to 1 indicate strong agreement, while values approaching -1 indicate strong disagreement.

Regarding the experiment on the PHD dataset, we utilized the widely adopted mean Average Precision (mAP) as a metric to evaluate the performance of our method, which is also applied in previous works \cite{bhattacharya2022show,wang2022pac,chen2021pr,rochan2020adaptive} in video highlight detection. We report the mAP on the test set and follow the way in \cite{wang2022pac} to calculate the mAP.
\subsection{Implementation Details}
% \subsubsection{Baseline:} Due to the lack of methods with similar objectives on the KLive dataset, we design variants of our proposed approach for ablation studies. 
\subsubsection{Baseline:} On KLive dataset, we compare our method with ID-based recommendation methods and several variants of our proposed approach as follows:
\begin{itemize}[leftmargin=*]
% \item \textbf{SIM \cite{pi2020search}:}
\item \textbf{W\&D \cite{cheng2016wide}:} This method trains wide linear models and deep learning network together to integrate the advantages of memorization and generalization in recommendation, but requires feature engineering in addition to raw features.
\item \textbf{DeepFM \cite{guo2017deepfm}:} This method utilizes the benefits of both factorization machines and deep learning by sharing input for the deep and wide components.
\item \textbf{DIFM \cite{lu2021dual}:} This approach leverages the benefit of input-aware factorization machines to address the issue of standard FMs that generate a fixed representation for varying input instances.
\item \textbf{DCN-M \cite{wang2021dcn}:} This approach is an enhanced version of DCN \cite{wang2017deep}, with greater effectiveness in learning both explicit and implicit feature crosses through the application of low-rank techniques to approximate such crosses.
\item \textbf{SelfAttention-plain:} We remove the ID embedding $E_u$ of the streamer of ContentCTR and replace the Perceiver block with a self-attention block.
\item \textbf{SelfAttention-input:} We replace the Perceiver block with a self-attention block and concatenate the ID embedding $E_u$ of the streamer with $E_f$ to be used as input for the self-attention. 
\item \textbf{CrossAttention-q:} We replace the Perceiver block with a cross-attention block. The query of every cross-attention layer is set as $E_u$, while the key and value are initialized with $E_f$. 
\item \textbf{CrossAttention-kv:} We replace the Perceiver block with a cross-attention block. The key and value of every cross-attention layer are set as $E_u$, while the query is initialized with $E_f$. 
\item \textbf{CrossAttention-qkv:} We replace the Perceiver block with a cross-attention block. The query is initialized with $E_u$, while the key and value are initialized with $E_f$.

% \item \textbf{DIN \cite{zhou2018deep}:} 4 DIN需要history_feature_list才能训练,没太懂什么是用户的历史行为
% \item \textbf{DIEN \cite{zhou2019deep}:} 5 DIEN 也需要history_feature_list才能训练，没太懂什么是用户的历史行为
% \item \textbf{MMOE \cite{ma2018modeling}:} 8会报错，需要多任务才能训练
\end{itemize}

\begin{figure*}[hbt]
\centering
\includegraphics[width=.95\textwidth]{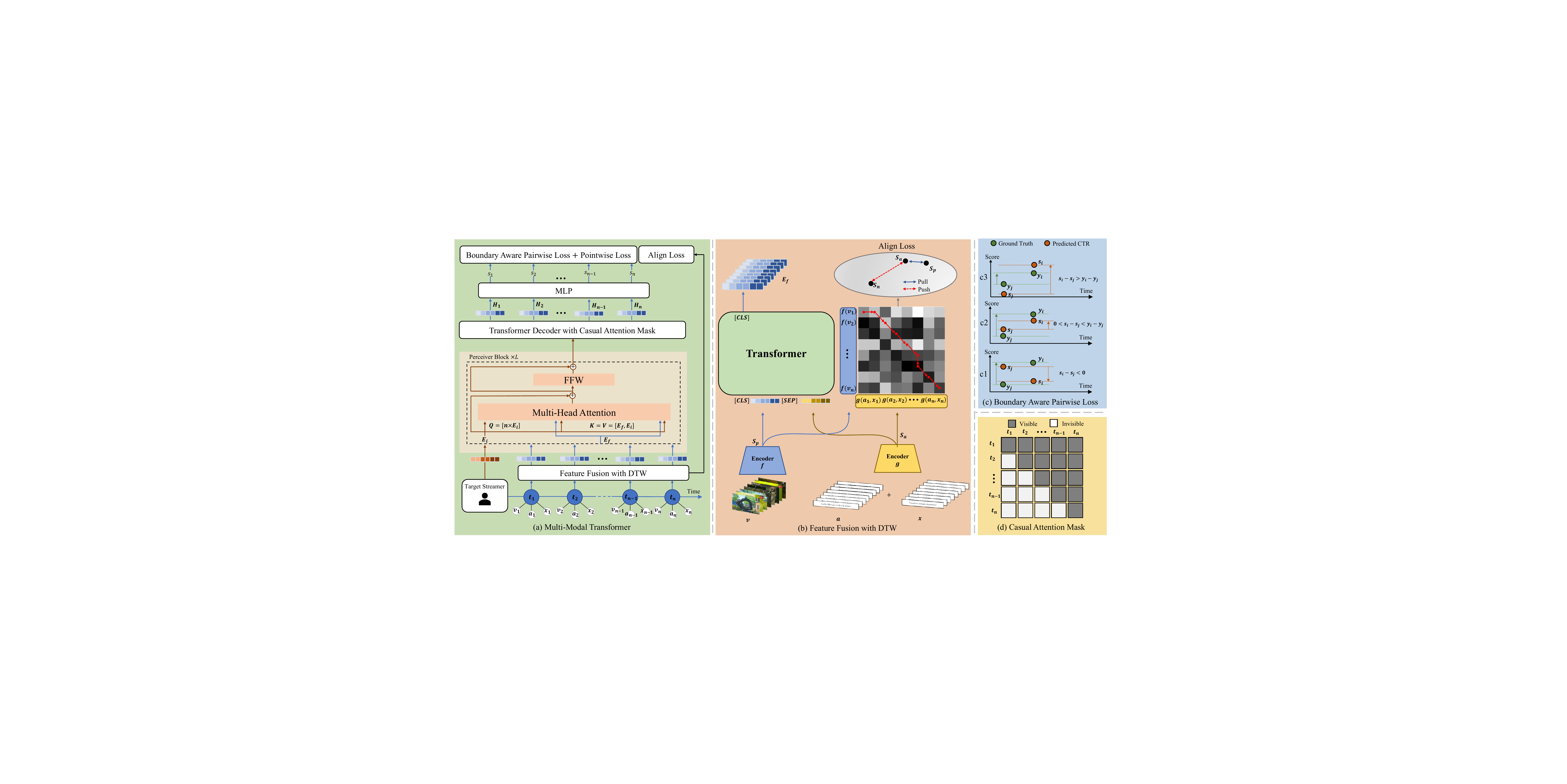}
\caption{\textbf{Visualization of the CTR prediction results for ContentCTR (red line), DIFM (green line) and ground truth CTR (grey line).}}
\label{fig6}
\vspace{-0.4cm}
\end{figure*}

On the PHD dataset, we make slight modifications to our proposed ContentCTR method and compare it with several classic baselines for personalized video highlight detection. Specifically, they are:
\begin{itemize}[leftmargin=*]
\item \textbf{Adaptive-H-FCSN} \cite{rochan2020adaptive}: This method employs a convolutional highlight detection network with a history encoder to learn user-specific highlight patterns.
\item \textbf{PR-Net} \cite{chen2021pr}: This method proposes a reasoning framework to explicitly learn frame-level patterns. It also employs contrastive learning to alleviate annotation ambiguity.
\item \textbf{PAC-Net} \cite{wang2022pac}: This method introduces a Decision Boundary Customizer (DBC) module and a Mini-History (Mi-Hi) mechanism to capture more fine-grained user-specific preferences.
\item \textbf{ShowMe} \cite{bhattacharya2022show}: This method leverages the content of both user history and target videos, using pre-trained features of YOLOv5 \cite{Jocher_YOLOv5_by_Ultralytics_2020}.
\item \textbf{ContentCTR-HL}: We modify our proposed ContentCTR model to adapt it to the  personalized video highlight detection task. First, we extract YOLOv5 features from images for training and testing. The text comes from the captions crawled from the YouTube video. Second, the ID embedding is replaced with the user history feature for personalized highlighting. Then, since there is no need to consider future information leakage in the video highlighting task, the causal attention mask is set to fully visible. Finally, when optimizing the network, only $\mathcal{L}_{Point}$ and $\mathcal{L}_{align}$ are reserved for optimization.
\end{itemize}
\subsubsection{Setup Detail:} During the training of ContentCTR on the KLive dataset, the layer number for the Pereciver block and Decoder block is set to 3, the input dimension $d=512$, the hidden dimension of the transformer $d_h$ is set to 64 and the number of attention head $n_h=8$. We utilize pre-trained swin as the Encoder $f(\cdot)$ and pre-trained Bert Chinese Large as the Encoder $g(\cdot)$. The swin and BERT are both pre-trained on image and text data from 39,000 streaming rooms. The tradeoff parameters, $\lambda_{1}$, $\lambda_{2}$, and $\lambda_{3}$, are respectively set to 0.65, 0.15, and 0.20. The $\sigma$ in $L_{Pair}^{1}$ is set to 10 and the number of negative sample $N$ is set to 8. We optimize ContentCTR for 12 epochs with a learning rate of $5 \times 10^{-5}$ using the Adam optimizer, and the global batch size is set to 48. During training, the gradient of swin and BERT feature extractors are open for fine-tuning, while the parameters of two MLP heads are updated, too. The training process takes about 34 hours on 8 Nvidia Tesla V100 GPUs. During the training and testing of ID-based recommendation methods on the KLive dataset, the input sparse features for these methods include the streamer item ID, live ID, timestamp ID, exposure count, comment count, gift count, click count, like count, follow count, room entry and exit count for each streaming segment. The objective of these methods is to regress the CTR of each segment. When training ContentCTR-HL on the PHD dataset, we first extract the YOLOv5 features for each frame of the historical highlight frames in the training and testing set. We use pre-trained YOLOv5 as the Encoder $f(\cdot)$ and pre-trained Bert English Large as the Encoder $g(\cdot)$. Then we train our network using the Adam optimizer with a batch size of 8 and an initial learning rate of $5 \times 10^{-5}$, with a cosine decay scheduler. The training is performed for 20 epochs on 8 NVIDIA Tesla V100 GPUs, which takes approximately 16 hours.

\subsection{Overall Performance Comparison}
%参考mPLUG-2: A Modularized Multi-modal Foundation Model  Across Text, Image and Video
\begin{table}[h]
    \centering
    \caption{
    Performances of different methods on KLive dataset
    }
    \label{overall}
    \vspace{-8pt}
    \setlength{\tabcolsep}{3.1pt}
    \scalebox{1.0}{
    \begin{tabular}{p{5.0cm}<{\raggedright}p{3.2cm}<{\centering}}
    \toprule
    Model  &  Tau $\tau$ \\
    \midrule
    \multicolumn{2}{l}{\textit{ID-based Recommendation Methods}}  \\
    \midrule
    \textbf{W\&D} \quad \quad \pub{DLRS'16}    & 0.5619 \\ %\up{0.xx\%}
    \textbf{DeepFM} \quad \pub{IJCAI'17}    & 0.5542  \\ %\up{0.xx\%}
    \textbf{DIFM} \quad \quad \pub{IJCAI'21}    & 0.5697 \\ %\up{0.xx\%}
    \textbf{DCN-M} \quad \quad \pub{WWW'21}    & 0.5629 \\ %\up{0.xx\%}
    \midrule
    \multicolumn{2}{l}{\textit{Variants of ContentCTR}} \\
    \midrule
    \textbf{SelfAttention-plain}   &  0.5868 \\
    \textbf{SelfAttention-input}   &   0.5718 \\ %\down{1.50\%}
    \textbf{CrossAttention-q}   &   0.5909 \\ %\up{0.41\%}
    \textbf{CrossAttention-kv}   &  0.5780 \\  %\down{0.88\%}
    \textbf{CrossAttention-qkv}   &    0.5883 \\ %\up{0.15\%}
    \midrule
    \cellcolor{Light}{\textbf{ContentCTR}}   &   \cellcolor{Light}{\textbf{0.5919} }\\ %\up{0.51\%}}
    \bottomrule
    \end{tabular}}
    \vspace{-10pt}
\end{table}
\subsubsection{Quantitative Results:} Table \ref{overall} summarizes the CTR prediction performances achieved by various methods on KLive dataset, indicating that \textbf{ContentCTR} surpasses all ID-based recommendation methods and the variations of \textbf{ContentCTR}. The findings reveal that the interaction approach among different modalities plays a crucial role in the final CTR prediction performance. Furthermore, compared to ID-based recommendation methods, ContentCTR demonstrates superior CTR prediction performance which suggests that the transformer based encoder-decoder architecture and Perceiver block is more efficient in capturing dynamic patterns.
\begin{table}[h]
    \centering
    \caption{Comparison with state-of-the-art alternative on PHD dataset.
    }
    \label{sotaphd}
    \vspace{-8pt}
    \setlength{\tabcolsep}{3.1pt}
    \scalebox{1.0}{
    \begin{tabular}{p{3.2cm}<{\raggedright}p{1.2cm}<{\centering}|p{1.5cm}<{\centering}|l}
    \toprule
    Model &  & Param.(M) & mAP(\%) \\
    \midrule
    \textbf{Adaptive-H-FCSN} \cite{rochan2020adaptive} & \pub{ECCV'20} & 197.35 & 15.65 \\
    \textbf{PR-Net} \cite{chen2021pr} & \pub{ICCV'21} & - & 18.66 \\
    \textbf{PAC-Net} \cite{wang2022pac} & \pub{ECCV'22} & 5.89 & 17.51 \\
    \textbf{ShowMe} \cite{bhattacharya2022show} & \pub{MM'22} & - & 16.40 \\
    \cellcolor{Light}{\textbf{ContentCTR-HL}} & \cellcolor{Light}{\pub{Ours}} & \cellcolor{Light}{66.35} & \cellcolor{Light}{\textbf{21.89}} \\
    \bottomrule
    \end{tabular}}
    \vspace{-8pt}
\end{table}

In order to answer \textbf{RQ5} and further evaluate the effectiveness of our method, we report the performance comparison of ContentCTR-HL with various state-of-the-art alternatives on PHD dataset. As shown in Table \ref{sotaphd}, our approach outperforms all other alternatives on PHD dataset, outperforming \textbf{PR-Net} \cite{chen2021pr} by +3.23\%. This result undoubtedly demonstrates the practicality and effectiveness of our method. The key improvement of our method can be derived from the following three aspects. Firstly, in contrast to other approaches that solely rely on visual features, our method leverages the multi-modal features of video frames and captions. Secondly, the sequence-to-sequence encoder-decoder architecture is proficient in detecting frame-level highlights. Thirdly, the incorporation of DTW alignment effectively alleviates the possible misalignments between captions and video frames.

\subsubsection{Qualitative Results:}
Moreover, we visualize several typical cases on KLive dataset in Figure \ref{fig6}. Figure \ref{fig6} illustrates the prediction results of \textbf{ContentCTR} in red line, \textbf{DIFM} in green line and the ground truth CTR in grey line. It is obvious that when the ground truth CTR arises abrupt fluctuations (shark rises or fails) due to the dynamic changing of content, \textbf{ContentCTR} is able to capture real-time content changes more effectively. However, \textbf{DIFM} tends to change more gradually, which is improper for high dynamic streaming highlight recommendation scenarios.

\subsection{Modality Impact(RQ1)}
We study the impact of different modalities on ContentCTR's performance on KLive dataset, as shown in Table \ref{modallity}. The results demonstrate that ContentCTR achieves a tau of $0.5919$ when all the modalities are engaged. We find that the visual modality has the most important impact on the model performance, causing a performance degradation of -4.72\% when removed. Apart from that, the text modality is the second most significant factor, leading to a -2.01\% degradation. Lastly, the ID embedding has the smallest but still significant effect on the model's performance, with a -0.51\% degradation when removed. This ablation study suggests that incorporating different modalities is essential for the model to predict CTR accurately. 
\begin{table}[h]
    \centering
    \caption{
    Ablation study on different modality impact.
    We study the effect of visual modality $v$, speech modality $a$, comment modality $x$ and item modality $u$.
    }
    \label{modallity}
    \vspace{-8pt}
    \setlength{\tabcolsep}{3.1pt}
    \scalebox{1.0}{
    \begin{tabular}{p{3.2cm}<{\raggedright}|p{0.5cm}<{\centering}p{0.5cm}<{\centering}p{0.5cm}<{\centering}p{0.5cm}<{\centering}|l}
    \toprule
    Model & $v$ & $a$ & $x$ & $u$ & Tau $\tau$ \\
    \midrule
    ContentCTR  & \cellcolor{Light}{\checkmark} & \cellcolor{Light}{\checkmark} & \cellcolor{Light}{\checkmark} & \cellcolor{Light}{\checkmark} & \cellcolor{Light}{\textbf{0.5919}} \\
    ContentCTR w/o item & \checkmark & \checkmark & \checkmark & - &  0.5868 \down{0.51\%} \\
     ContentCTR w/o text & \checkmark & - & - & \checkmark &  0.5718 \down{2.01\%} \\
     ContentCTR w/o visual & - & \checkmark & \checkmark & \checkmark & 0.5447 \down{4.72\%} \\
    \bottomrule
    \end{tabular}}
    \vspace{-10pt}
\end{table}

\subsection{Perceiver Block(RQ2)}
% In order to verify the superiority of the Perceiver block, we designed several variants of our methods on KLive dataset. These variants are as follows: \textbf{Baseline:} We removed the ID embedding $E_u$ of the streamer and replaced the Perceiver block with a self-attention block. \textbf{Model1:} We replaced the Perceiver block with a self-attention block and concatenated the ID embedding $E_u$ of the streamer with $E_f$ to be used as input for the self-attention. \textbf{Model2:} We replaced the Perceiver block with a cross-attention block. The query of every cross-attention layer was set as $E_u$, while the key and value were initialized with $E_f$. \textbf{Model3:} We replaced the Perceiver block with a cross-attention block. The key and value of every cross-attention layer were set as $E_u$, while the query was initialized with $E_f$. \textbf{Model4:} We replaced the Perceiver block with a cross-attention block. The query was initialized with $E_u$, while the key and value were initialized with $E_f$.

As demonstrated in Table \ref{overall}, the Perceiver block surpasses all the variants of ContentCTR in terms of performance. Table \ref{overall} provides the following insights.

Firstly, the results indicate that inputting the ID embedding directly into the transformer model fails to fully leverage personalized information. The self-attention architecture is not suitable for detecting the correlation between different modalities.

Secondly, concerning cross-attention architectures, it is critical to carefully design the interaction approach between the ID embedding and other content-based modalities. The traditional query-key-value-based cross-attention mechanism proves to be less effective than the Perceiver block. 

\subsection{DTW Alignment and Pairwise Loss(RQ3\&RQ4)}
% We design \textbf{Model5-Model9} to investigate the effect of different loss function with $L_{Point}$, $L_{Pair}^{0}$, $L_{Pair}^{1}$, $L_{Pair}^{2}$, $L_{Pair}^{3}$ and $L_{align}$. As shown in Table \ref{lossfunc}, when optimize \textbf{Model8} with $L_{Point}$ and $L_{Pair}^{2}$, \textbf{Model8} suffer great performance degradation with a 5.05\% degradation. We hyposize that when only optimized in Part 1, we gradient changeing is too big which is harmful to model the study the contrastive information of highlight frames and no- highlight frames. However, when \textbf{Model9} only optimize on Part 2，it gains a 0.66\% of performance improvement but is still not outstanding to \textbf{Model6} which joint optimized on Part 1 and Part 2 which bring a significant improvement with 1.11 \%. As for \textbf{Model6}, although it also gain some performance improvement, but when training, we find the pointwise loss suffer from the over optimize from part 3 which is shown in Figure 4

We design \textbf{Model1-Model5} to investigate the impact of different loss functions on KLive dataset, which include $L_{Point}$, $L_{Pair}^{0}$, $L_{Pair}^{1}$, $L_{Pair}^{2}$, $L_{Pair}^{3}$, and $L_{align}$. According to Table \ref{lossfunc}, when we optimized \textbf{Model4} with $L_{Point}$ and $L_{Pair}^{2}$, its performance significantly degraded by -5.05\%. We hypothesize that the gradient changes are too drastic when optimizing only Part 1, which is detrimental to modeling the contrastive information of highlight frames and non-highlight frames. However, when optimizing only Part 2 with $L_{Pair}^{3}$, \textbf{Model5} showed a performance improvement of +0.66\%, but it still falls behind \textbf{Model3}, which is jointly optimized on Part 1 and Part 2 with $L_{Pair}^{1}$, resulting in a significant improvement of +1.11\%.
\begin{table}[h]
\vspace{-5pt}
    \centering
    \caption{
    Ablation study of ContentCTR on pairwise loss and DTW align loss.
    We study the effect of loss function $L_{Point}$, $L_{Pair}^{0}$, $L_{Pair}^{1}$, $L_{Pair}^{2}$, $L_{Pair}^{3}$ and $L_{align}$.
    }
    \label{lossfunc}
    \vspace{-8pt}
    \setlength{\tabcolsep}{3.1pt}
    \scalebox{0.85}{
    \begin{tabular}{p{1cm}<{\centering}|p{0.8cm}<{\centering}p{0.6cm}<{\centering}p{0.6cm}<{\centering}p{0.6cm}<{\centering}p{0.6cm}<{\centering}p{0.7cm}<{\centering}|ll}
    \toprule
    Model & $L_{Point}$ & $L_{Pair}^{0}$ & $L_{Pair}^{1}$ & $L_{Pair}^{2}$ & $L_{Pair}^{3}$ & $L_{align}$ & Tau $\tau$  & \gc{avg. $s/y$}\\
    \midrule
   \textbf{Model1} & \checkmark & - & - & - & - & - & 0.5761  & \gc{1.2789}\\
\textbf{Model2} & \checkmark & \checkmark & - & - & - & - & 0.5857 \up{0.96\%} & \gc{1.3164}  \\
\textbf{Model3} & \checkmark & - & \checkmark & - & - & - & 0.5872 \up{1.11\%} &\gc{1.3021} \\
\textbf{Model4} & \checkmark & - & - & \checkmark & - & -  &  0.5256  \down{5.05\%} &\gc{1.3414}   \\
\textbf{Model5} & \checkmark & - & - & - & \checkmark & -  &  0.5824 \up{0.66\%} &\gc{1.2194}  \\
\cellcolor{Light}{\textbf{Ours}} & \cellcolor{Light}{\checkmark} & \cellcolor{Light}{-} & \cellcolor{Light}{\checkmark}  & \cellcolor{Light}{-} & \cellcolor{Light}{-} & \cellcolor{Light}{\checkmark}  &  \cellcolor{Light}{\textbf{0.5919} \up{1.58\%}} &\cellcolor{Light}{\gc{1.3011}}  \\
    \bottomrule
    \end{tabular}}
    \vspace{-5pt}
\end{table}
\begin{figure}[h]
\centering
\includegraphics[width=.45\textwidth]{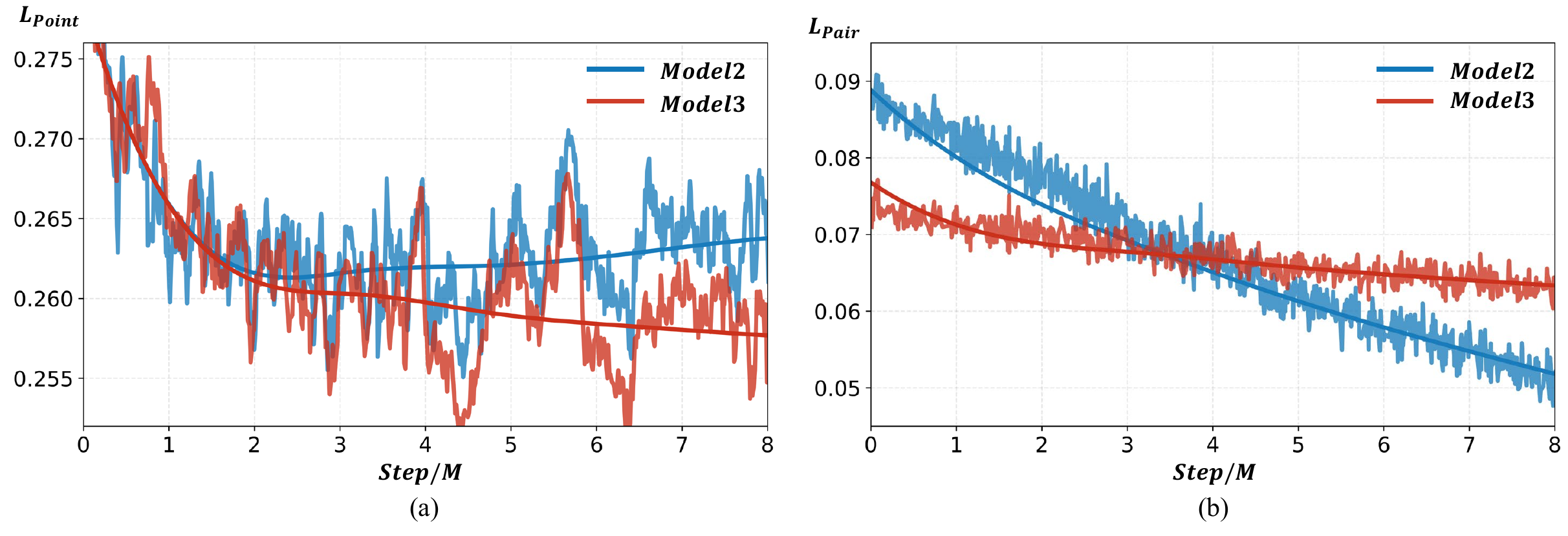}
\caption{\textbf{The change of $L_{Point}$ and $L_{Pair}$ of \textbf{Model2} (blue line) and \textbf{Model3} (red line) during training.} It is obvious that the pairwise loss of \textbf{Model2} is over-optimized which causes the pointwise loss to collapse in \textbf{Model2} while both losses remain normal in \textbf{Model3}.}
\label{fig4}
\vspace{-0.2cm}
\end{figure}
Although \textbf{Model2} with $L_{Pair}^{0}$ has also shown performance improvement, we have noticed that during training, the pointwise loss tends to collapse due to the over-optimization of pairwise loss on Part 3, which is shown in Figure \ref{fig4}. 

\begin{figure*}[hbt]
\centering
\includegraphics[width=.95\textwidth]{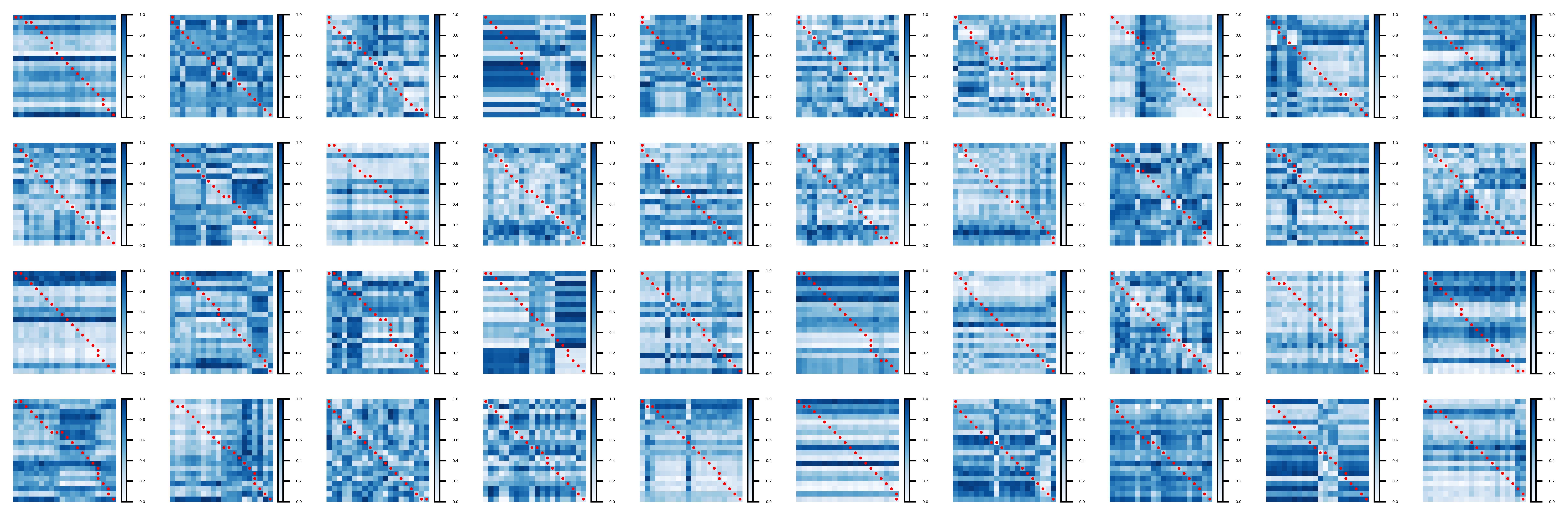}
\caption{\textbf{Visualization of Dynamic Time Warping (DTW) alignment results for the KLive dataset. Each subfigure represents the cosine similarity between video and text feature sequences. The red point denotes the DTW alignment path found by our algorithm.}}
\label{fig5}
\vspace{-0.5cm}
\end{figure*}

As expected, when \textbf{Model3} incorporates $L_{align}$ for additional optimization, the DTW alignment loss shows further improvement by +0.47\%. Figure \ref{fig5} shows some visualizations of DTW alignment results for the KLive dataset, which reveal that misalignment cases do exist in streaming scenarios. This also indicates that the involvement of $L_{align}$ helps ContentCTR in training better visual and text encoders which reduces the possible misalignment between the two.

\subsection{Ablation Study on PHD}
We also conducted an additional ablation study on the PHD dataset based on ContentCTR-HL, analyzing the impact of highlight frames, captions, DTW align loss and input frames number. As shown in Table \ref{contencHLabl}, the performance of ContentCTR-HL decreased by -2.34\% when the history feature is removed, indicating that the history feature of highlight frames is crucial for personalized video highlight detection. Similarly, the absence of captions resulted in a -1.11\% decrease in ContentCTR-HL's performance, highlighting the usefulness of text modality for the model's detection capability. Interestingly, ContentCTR-HL is found to be less sensitive to DTW alignment loss. We hypothesize that this phenomenon is caused because most captions generated by the official YouTuber are accurate enough. Apart from that, since $L_{align}$ is a self-learning contrastive loss, it may require a larger data scale to take effect, but currently, only 30\% of all videos in PHD have downloadable captions. We also find that the performance of ContentCTR-HL improves with the increase of the input frames number. However, the increase in performance is insignificant when the number of frames is increased from 96 to 192.
\begin{table}[h]
    \centering
    \caption{Ablation Study on ContentCTR-HL.
    }
    \label{contencHLabl}
    \vspace{-8pt}
    \setlength{\tabcolsep}{3.1pt}
    \scalebox{0.85}{
    % \begin{tabular}{p{3.2cm}<{\raggedright}p{1.2cm}<{\centering}|p{1.5cm}
    \begin{tabular}{p{3.6cm}<{\raggedright}|p{0.7cm}<{\centering}p{0.7cm}<{\centering}p{0.7cm}<{\centering}p{2cm}<{\raggedright}}
    \toprule
    % Model &   $n=32$ & $n=64$ & $n=96$  & $n=192$\\
    \multicolumn{1}{c|}{\multirow{2}{*}{Model}} & \multicolumn{4}{c}{Frames Number $n$}                 \\ \cline{2-5} 
\multicolumn{1}{c|}{}                       & 32  & 64 & 96 & 192 \\ \midrule
    % \midrule
    % \cline{4-6}
    ContentCTR-HL(Ours) & 18.29 & 19.54 & 21.59 & 21.89 \\
    Ours w/o history & - & - & - & 19.55 \down{2.34\%} \\
    Ours w/o caption & - & - & - & 20.06  \down{1.11\%} \\
    Ours w/o DTW & - & - & - &21.75 \down{0.14\%}\\
    \bottomrule
    \end{tabular}}
    \vspace{-7pt}
\end{table}

\subsection{Online Experiments}
%%% 不清楚怎么写线上实验的部分，下面是一篇文章的范例
% We perform a careful online A/B testing on Taobao from 2020-11 to 2020-12, which is under the bucket tests. One bucket is selected for baseline and another bucket for our model. Each bucket serves about 0.5 million users per day. For the whole bucket, the goal is to increase the user stickiness and activity. During nearly two months of A/B testing, our approach contributes up to 9.32\% number of browsing videos per user, 10.45\% dwell time per user, and 12.10\% complete-watch ratio per video, compared with the online baseline (i.e., Cross-BST). These online benefits from our method are crucial for the micro-video commercial layout of Taobao. For example, advertising videos will have more broadcast opportunities with the increased number of users watching micro-videos. The proposed method has already been fully deployed online in December 2020 and serves over 30 million users on mobile Taobao every day. The details of the online deployment are introduced in Appendix A.4.
We test the proposed framework in real-world live streaming scenarios through online A/B testing. The experiment is conducted over four consecutive days, with traffic randomly assigned to either the baseline method or our method. In the baseline group, candidate live rooms are sorted by scores produced by a traditional recommendation model, e.g., Click-Through Rate $s_{ctr}$, Long-View-Through Rate $s_{lvtr}$. Note that these scores focus on capturing the long-term relationship between streamers and users. In contrast, our method utilizes content-based CTR as an additional factor like $s_{exp}$. This term is capable of catching the highlight moment and show the most attractive live contents to users. The results show that our method achieve a 2.9\% and  5.9\% improvements in terms of CTR and live play duration, respectively, which demonstrate the effectiveness of our content based model.
\begin{equation}
\begin{aligned}
s_{base} & = s_{ctr} + s_{lvtr} + ... \\
s_{exp} & = s_{ctr} + s_{lvtr} + ... + s_{ContentCTR},
\end{aligned}
\end{equation}
\label{abtest}

% Moreover, we visualize several typical cases in Figure \ref{} in order to demonstrate that our ContentCTR is able to capturing real-time content changes.

\section{Conclusion}
In this paper, we study the task of Click-Through Rates prediction in live streaming scenario. We introduce ContentCTR, which utilizes a multimodal transformer to achieve frame-level CTR prediction. Specifically, we propose a streamer-personalized Perceiver Block that fuses ID embedding, visual, audio, and comment embedding. The decoder network outputs the final CTR prediction for each frame. To address the possible misalignment between video frames and texts, we carefully design a Dynamic Time Warping (DTW) alignment loss for optimization. Additionally, the boundary-aware constrained pairwise loss demonstrates better performance when combined with the pointwise loss. We conduct comprehensive experiments on both the KLive dataset and the public PHD dataset, which demonstrate the effectiveness of our methods in both streaming CTR prediction and video highlight detection tasks, compared with state-of-the-art methods. Moreover, the proposed method has been deployed online on the company's short video platform and serves over three million daily users.


% \begin{acks}
% \end{acks}

% \bibliographystyle{ACM-Reference-Format}
% \bibliography{cikm}

\appendix








\end{document}